\documentclass{article}
\usepackage{spconf,amsmath,epsfig}
\usepackage{graphicx}
\usepackage{multirow}
\usepackage{amsmath}
\usepackage{float}
\usepackage{epstopdf}
\usepackage{tabularx}
\usepackage{subfig}
\usepackage{caption}
\begin{document}\sloppy
\newcolumntype{C}[1]{>{\centering\arraybackslash}p{#1}}

\title{Cross Domain Knowledge Transfer for Unsupervised Vehicle Re-identification}
%
\name{Jinjia Peng, Huibing Wang, Tongtong Zhao and Xianping Fu}
\address{Information Science and Technology College, Dalian Maritime University, Dalian 116026, China}

\maketitle

\begin{abstract}
Vehicle re-identification (reID) is to identify a target vehicle in different cameras with non-overlapping views. When deploy the well-trained model to a new dataset directly, there is a severe performance drop because of differences among datasets named domain bias. To address this problem, this paper proposes an domain adaptation framework which contains an image-to-image translation network named vehicle transfer generative adversarial network (VTGAN) and an attention-based feature learning network (ATTNet). VTGAN could make images from the source domain (well-labeled) have the style of target domain (unlabeled) and preserve identity information of source domain. To further improve the domain adaptation ability for various backgrounds, ATTNet is proposed to train generated images with the attention structure for vehicle reID. Comprehensive experimental results clearly demonstrate that our method achieves excellent performance on VehicleID dataset.

\end{abstract}
\section{Introduction}
\begin{figure}
\centering
\includegraphics[width=8.5cm]{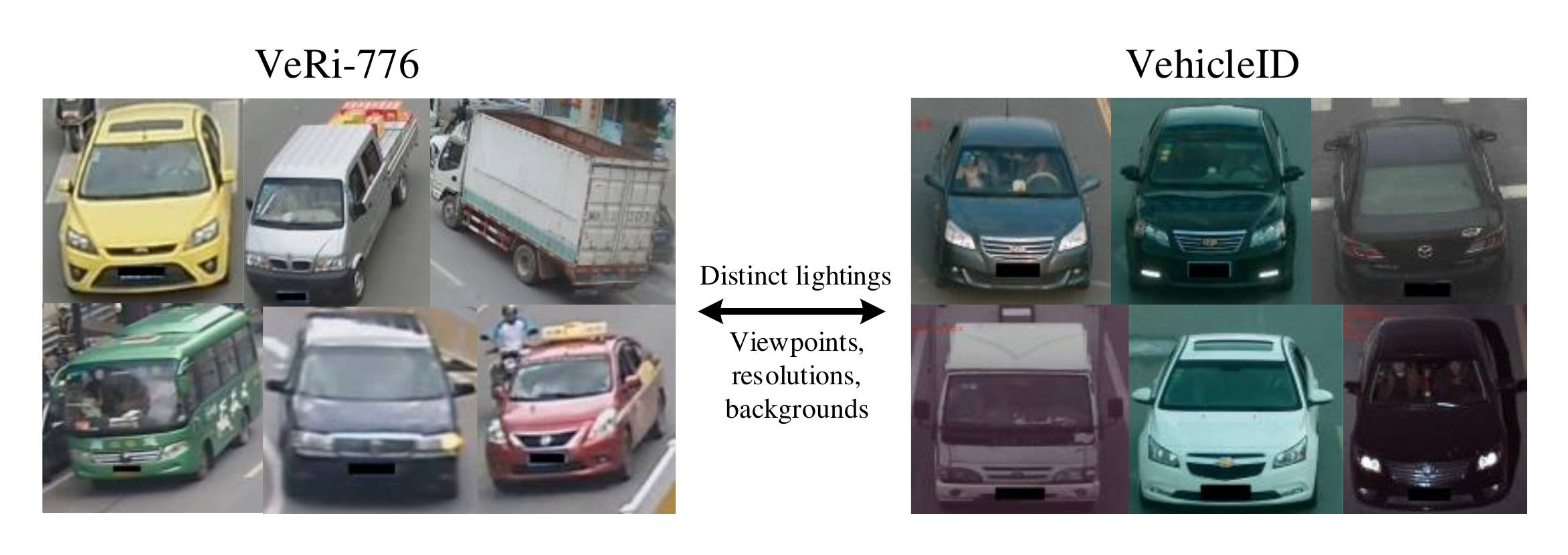}
\caption{Illustration of the datasets bias between VeRi-776 and VehicleID. The VeRi-776 and VehicleID present different styles,e.g., distinct lightings, backgrounds, viewpoints, resolutions etc.} \label{fig1}
\end{figure}

Video surveillance for traffic control and security plays a significant role in current public transportation systems. The task of vehicle reID often undergoes intensive changes in appearance and background. Captured images in different datasets by different cameras is a primary cause of such variations. Usually, datasets differ form each other regarding lightings, viewpoints and backgrounds, even the resolution, etc. As shown in Fig.\ref{fig1}, images in VeRi-776 are brighter and have more viewpoints than images in VehicleID. And images in VehicleID have higher resolution than images in VeRi-776. Besides that, it could not contain all cases in real scenario for every dataset, which makes different datasets form their own unique style and causes the domain bias among datsets. For reID \cite{wu20193D}, most existing works follow the supervised learning paradigm which always trains the reID model using the images in the target dataset first to adapt the style of the target dataset \cite{wang2017effective}\cite{Wu2018}\cite{wang2015robust}\cite{wu2018deep}. However, it is observed that, when the well-trained reID model is tested on other dataset without fine-tuning, there is always a severe performance drop due to the domain bias.

There are few studies on vehicle reID about the cross-domain adaptation. And only a few methods exploit unlabeled target data for unsupervised person reID modelling \cite{DengW}\cite{WangJ}\cite{WeiL}. However, some of them \cite{wu2019cycle} need extra information about source domain while training, such as attribute labels and spatio-temporal labels, which are not existing on some datasets. And there are only several methods exploiting unsupervised learning \cite{Wang2016Iterative,AMultiview} without any labels, for instance, SPGAN \cite{DengW} and PTGAN \cite{WeiL}. SPGAN is designed for person reID that integrates a SiaNet with CycleGAN \cite{KimT} and it does not need any additional labels during training. However, though SPGAN is effective on the person transfer task, it causes deformation and color distortion in vehicle transfer task in our experiment. PTGAN is composed of PSPNet \cite{ZhaoH} and CycleGAN to learn the style of target domain and maintain the identity information of source domain. In order to keep the identity information, PSPNet is utilized to segment the person images first. As we all know, it needs pre-trained segment model for PSPNet, which increases the complexity of the training stage.

To sump up, this paper proposes an end-to-end image-to-image translation network for the vehicle datasets, which named VTGAN. To preserve the identity information of images from source domain and learn style of images from target domain, for every generator in VTGAN, it is composed of a content encoder, a style encoder and a decoder. An attention model is proposed in the content encoder to preserve the identity information from the source domain. And the style encoder is designed to learn the style of the target domain with the style loss. Furthermore, VTGAN does not need any labels and paired images during the translation procedure, which is closer to the real scenario. To better adapt the target domain (unlabeled), ATTNet is designed for vehicle reID with the generated images obtained from the stage of translation. It has better generalization ability through the proposed attention structure to focus on the foreground and neglect the background information of the input image as much as possible during training procedure. In summary, our contributions can be summarized into two aspects:

1) We propose VTGAN to generate the images which have the style of target domain and preserve identity information of source domain. It is an efficient unsupervised learning model and works by transferring content and style between different domains separately.

2) ATTNet is presented to train the generated images, which is based on attention structure and could extract more distinctive cues while suppressing background for vehicle reID task.

\section{Method}
\subsection{Overview}

Our ultimate goal is to perform vehicle reID model in an unknown target domain for which are not labeled directly. Hence, we introduce a two-step vehicle reID method based on Generative Adversarial Network (GAN). The first step is to transfer the style between source domain and target domain. In this step, the VTGAN is proposed to generate images which have the style of target domain and preserve the identity information of source domain. After generating the style transferred images, in the second step, we design a multi-task network with the attention structure to obtain more discriminative features for vehicle reID.

\subsection{VTGAN}

\begin{figure}[ht]
\centering
\includegraphics[width=8.5cm]{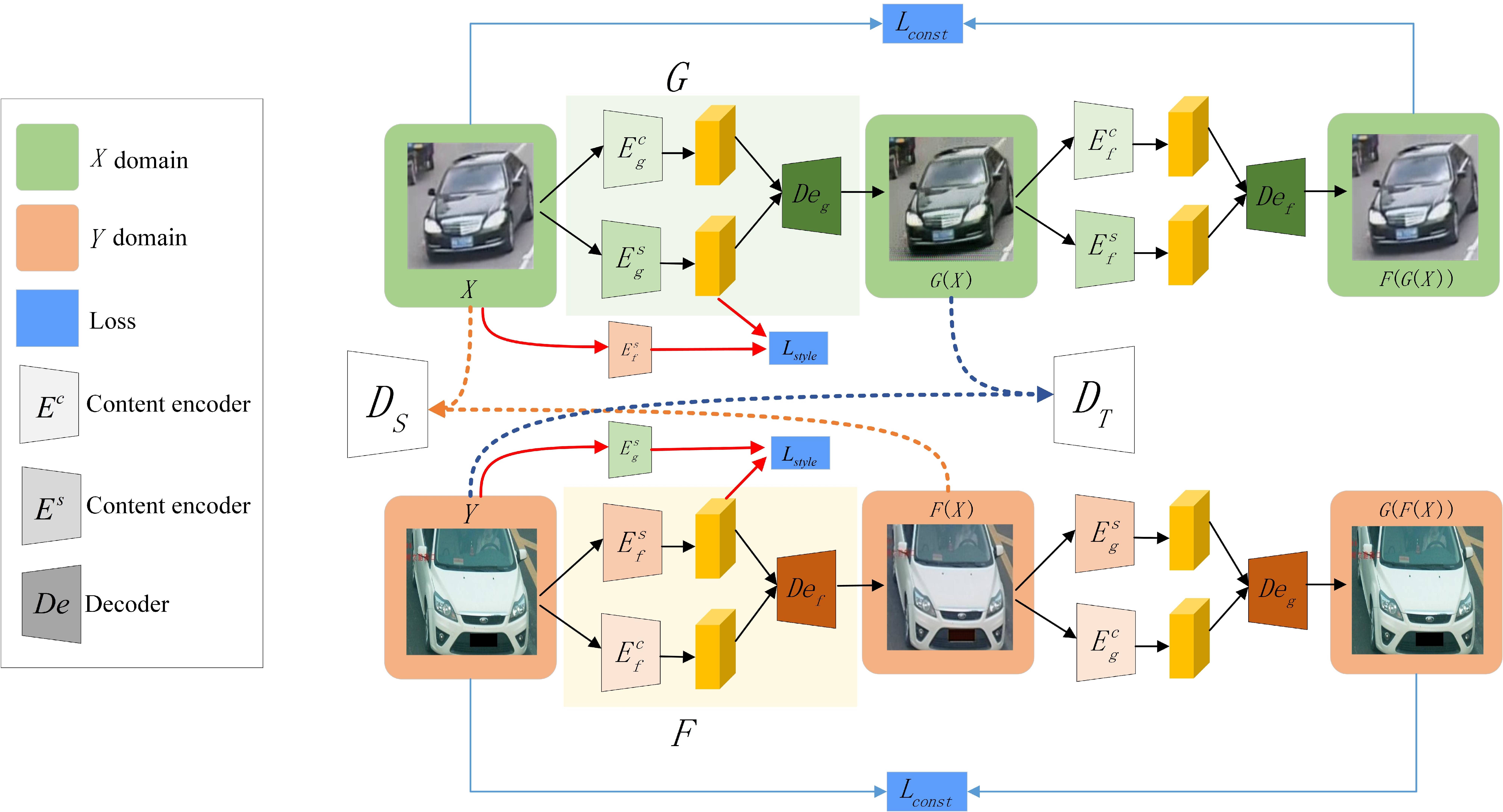}
\caption{The structure of VTGAN. VTGAN contains two mapping functions: $G:X{\rightarrow}Y$ and $F:Y{\rightarrow}X$, and associated adversarial discriminators $D_{T}$ and $D_{S}$. $L_{const}$ and $L_{style}$ represents cycle consistency loss and gram loss which are employed to further regularize the mappings (best viewed in color).} \label{fig2}
\end{figure}
VTGAN is designed to transfer the style between source domain and target domain in the case of preserving the identity information of images from source domain. As illustrated in Fig.\ref{fig2}, VTGAN consists of generators ${G,F}$, and domain discriminators ${D_{S}, D_{T}}$ for both domains. For each generator in VTGAN, it contains content encoder ${E^c}$, style encoder ${E^s}$ and decoder ${De}$ three components. ${E^c}$ is designed to preserve the identity information from images of source domain through the proposed attention model, which could extract the foreground while suppressing background. And to learn the style of target domain, the ${E^s}$ with the gram loss is added to the translation network. At last, the decoder $De$ embeds the output of ${E^c}$ and $E^s$ to generate the translated image.

\subsubsection{Content Encoder}

In order to keep the identity information from source domain, the attention model is designed to assign higher scores of visual attention to the region of interest while suppressing background in the content encoder.

\begin{figure}[ht]
\centering
\includegraphics[width=7cm]{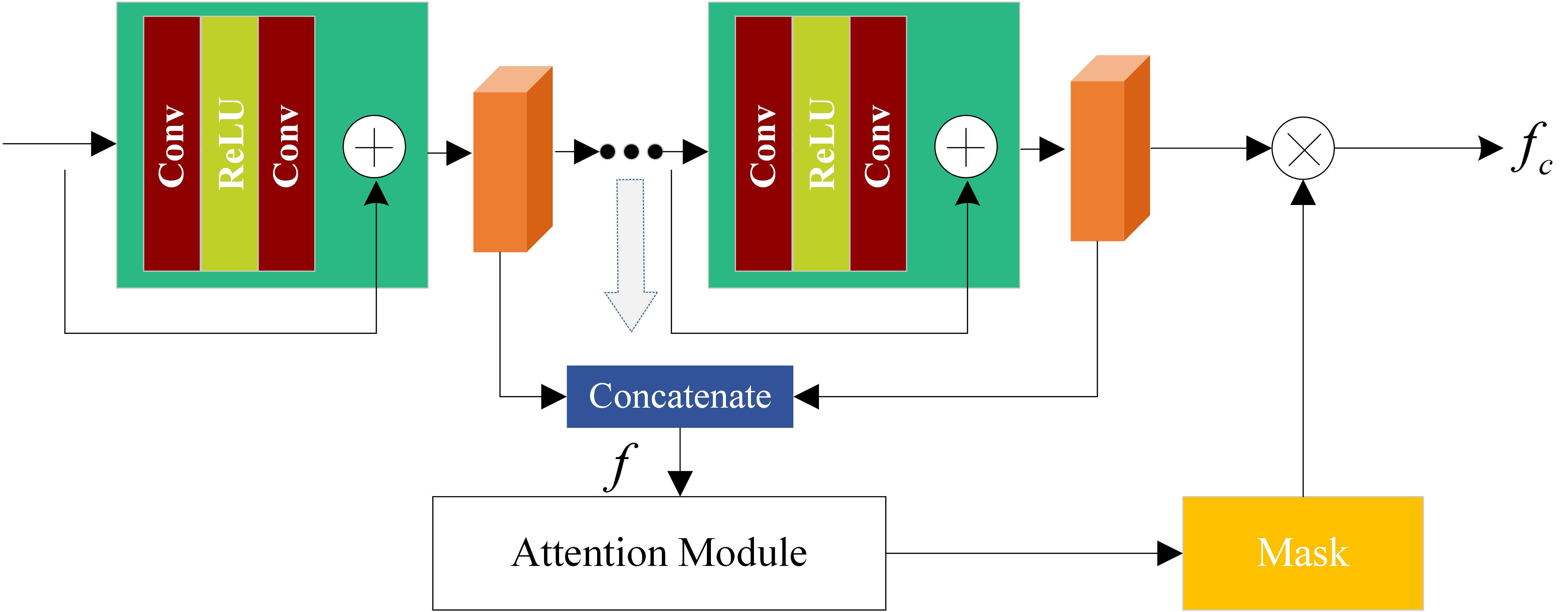}
\caption{The illustration of proposed attention structure.} \label{fig3}
\end{figure}

As shown in Fig.\ref{fig3}, we denote the input feature map of attention model as $f$. In this work, a simple feature fusion structure is utilized to generate the $f$. We fuse the every output of the ResBlock to form $f$, which can be formulated as $f=[f_{r1}, f_{r2}, ..., f_{r9}]$, where $f_{ri}$ is the $ith$ feature map generated by the $ith$ ResBlock. $i\in [1,9]$ and $[\cdot]$ denotes the concatenation operation. For the feature vector $f_{i,j}\in{\Re^C}$ of the feature map at the spatial location $(i,j)$, we can calculate its corresponding attention mask $a_{i,j}$ by

\begin{equation}
a_{i,j}=Sigmod(FC(f_{i,j};W_{a}))
\end{equation}
where $FC$ is the Full Connected layer (FC) to learn a mapping function in the attention module and $W_{a}$ are the weights of the FC. The final attention mask $\alpha=[a_{i,j}]$ is a probability map obtained using a Sigmod layer. The scores represent the probability of foreground in the input image.

And after the attention model, a mask $a$ is generated, which has high scores for foreground and low scores for background. Hence, the attended feature map $f_{c}$ is computed by element-wise product of the attention mask and the input feature map, which could be described as follows:

\begin{equation}
f_{c(i,j)}=a_{i,j}\otimes{f_{i,j}}
\end{equation}
where $(i,j)$ is the spatial location in mask $a$ or feature map $f_{c}$. And $\otimes$ is performed in an element-wise product.


\subsubsection{Style Encoder}
Besides the content branch, there is a branch to learn style of target domain. In this branch, different with the $E^c_g$ and $E^c_f$, the style network $E^s_g$ and $E^s_f$ do not contain the attention model. To learn the style of the target domain, $E^s_g$ is designed with the gram loss to output the style features $f_{s}$ which has similar distribution of the target domain $Y$. The gram loss could be formulated as follows:

\begin{equation}
\begin{split}
L_{style}=  {\frac{1}{NM}{{(T(x)-A(y))^2}}}+{\frac{1}{NM}{{(T(y)-A(x))^2}}}
\end{split}
\end{equation}
where $N$ is the number of feature maps, $M$ is calculated by $width\times{height}$, $width$ and $height$ represent the width and height of images. $T(x)$, $T(y)$, $A(y)$ and $A(x)$ are the gram matrix of output features $E_{g}^s(x)$, $E_{f}^s(y)$, $E_{g}^s(y)$ and $E_{f}^s(x)$, respectively.

\subsubsection{Decoder Network}
For the decoder network, it is composed of two deconvolution layers and a convolutional layer to output the generated images $G(I)$. The input of the decoder network is the combination of $f_{c}$ and $f_{s}$ which represent the content features and style features, respectively. In this paper, we employ a concatenate layer to fuse $f_{c}$ and $f_{s}$ and a global skip connection structure to make training faster and resulting model generalizes better, which could be expressed as:

\begin{equation}
G(I)=tanh(conv(deconv(deconv([f_{c}, f_{s}])+f_{e2})))
\end{equation}
where $[.]$ represents the concatenate layer. And $f_{e2}$ represents the feature map generated by the second stride convolution blocks. $f_{c}$ and $f_{s}$ are the output of content encoder and style encoder, respectively.

\subsubsection{Loss function}
We formulate the loss function in VTGAN as a combination of adversarial loss, content loss and style loss:
\begin{equation}
L=L_{GAN}+\lambda_{1}L_{id}+\lambda_{2}L_{style}
\end{equation}
where the $\lambda_{1}$ and $\lambda_{2}$ control the relative importance of three objectives. The style loss $L_{style}$ could be calculated by Eq.(3). VTGAN utilizes the target domain identity constraint as an auxiliary for image-image translation. Target domain identity constraint was introduced by \cite{TaigmanY} to regularize the generator to be the identity matrix on samples from target domain, written as:
\begin{equation}
\begin{split}
L_{id}=  E_{y\sim p_{data}(y)}||F(y)- y||_1  +E_{x\sim p_{data}(x)}||G(x)- x||_1
\end{split}
\end{equation}

For $L_{GAN}$, it consists of three parts which two adversarial losses and a cycle consistency loss. VTGAN applies adversarial losses to both mapping functions. For the generator $F$ and its discriminator $D_{T}$, the objective could be expressed as:
\begin{equation}
\begin{split}
L_{T}(F,D_T,X,Y)= & E_{x\sim p_{data}(x)}[(D_T(x))^2] \\ & + E_{y\sim p_{data}(y)}[||D_{T}(F(y))-1||_1]
\end{split}
\end{equation}
where, $X$ and $Y$ represent the source domain and target domain, respectively. $p_{data}(x)$ and $p_{data}(y)$ denote the sample distributions in the source and target domain. The objective of generator $G$ and discriminator $D_{S}$ also could be built. Besides, the VTGAN requires $F(G(x))\approx x$ and $G(F(y))\approx y$ when it learns the mapping of $F$ and $G$. So the cycle consistency loss is employed in VTGAN which could make the network more stable. The cycle consistency loss could be defined as:
\begin{equation}
\begin{split}
L_{cyc}(F,G,X,Y)= & E_{x\sim p_{data}(x)}[||F(G(x))- x||_1] \\ & +E_{y\sim p_{data}(y)}[||G(F(y))- y||_1]
\end{split}
\end{equation}

\subsection{ATTNet}
As we all know, the diversity background is a big factor for the problem of cross domain. And in order to make the reID model adapt to the target domain, we are facing a condition that it is better to focus on the vehicle images and neglect the background when we train the feature learning model. Hence, a two-stream reID network with attention structure is designed in this paper.

\begin{figure*}
\center
\includegraphics[width=13cm]{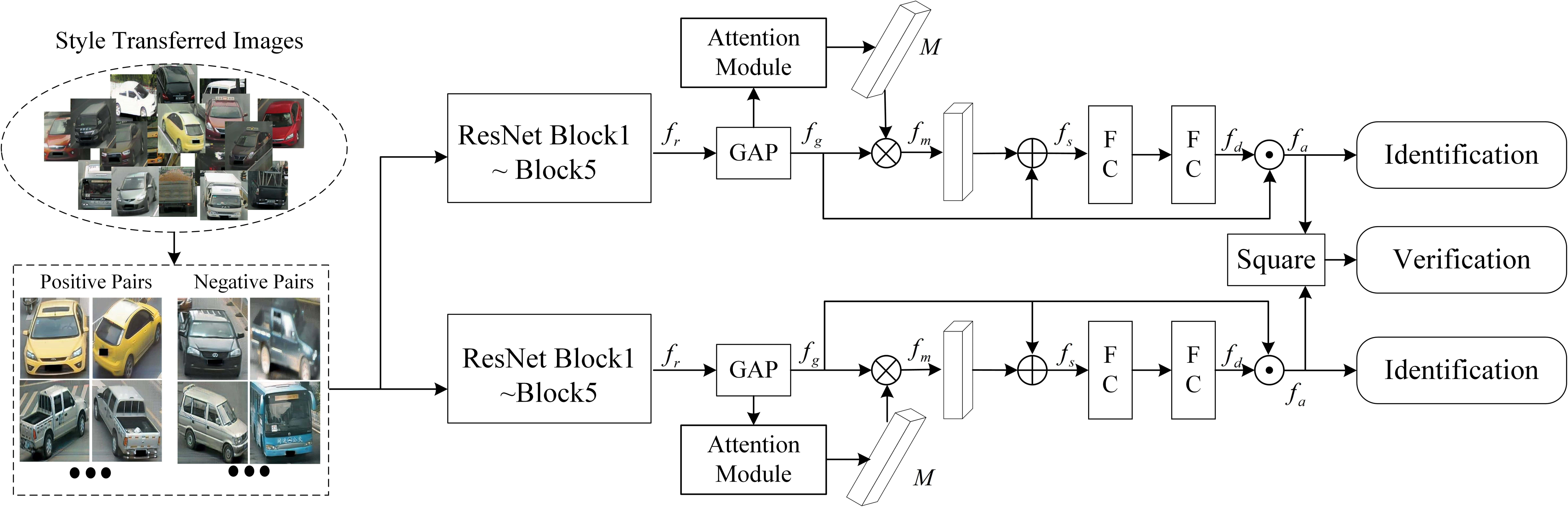}
\caption{The structure of ATTNet.} \label{fig4}
\center
\end{figure*}
As shown in Fig.\ref{fig4}, the input images are obtained from the image generation module and are divided into positive and negative sample pairs. For one branch, the input image is fed into five ResNet Blocks \cite{HeK} to output the feature maps $f_{r}$ with the size of $7\times 7\times 2048$. Then they are passed into a Global Average Pooling (GAP) layer to obtain the feature map $f_{g}$. $f_{g}$ is utilized to generate the mask $M$ through the proposed attention structure. Given the feature map $f_{r}$, its attention map is computed as $M = Softmax(Conv(f_{g}))$, where the one $Conv$ operator is $1\times 1$ convolution. After obtaining the attention map $M$, the attended feature map could be calculated by $f_m = f_{g}\otimes M$. The operator $\otimes$ is performed in an element-wise product. Then the attended feature map $f_{m}$ will be fed into the subsequent structure. A shortcut connection architecture is introduced to embed the input of the attention network directly to its output with an element-wise sum layer, which could be described as $f_s = f_{g} + f_{m}$. In this way, both the original feature map and the attended feature map are combined to form features $f_{a}$ and utilized as input to the subsequent structure. And after two FC layers, we could obtain the feature $f_{d}$. At last, a skip connection structure is utilized to integrate $f_{g}$ and $f_{d}$ by the concatenate layer to obtain more discriminative features for identification task and verification task, which could be described as $f_{a} = [f_{d}, f_{g}]$.

\section{Experiments}

\subsection{Datasets}

VeRi-776 \cite{LiuX'} is a large-scale urban surveillance vehicle dataset for reID. This dataset contains over 50,000 images of 776 vehicles with identity annotations, camera geo-locations, image timestamps, vehicle types and colors information. In this paper, 37,781 images of 576 vehicles are employed as a train set. VehicleID \cite{LiuH} is a surveillance dataset from real-world, which contains 26267 vehicles and 221763 images in total. From the original testing data, four subsets, which contain 800, 1600, 2400 and 3200 vehicles, are extracted for vehicle search in different scales.

\subsection{Implementation Details}
For VTGAN, we train the model in the tensorflow \cite{AbadiM} and the learning rate is set to 0.0002. Note that, we do not utilize any label notation during the learning procedure. The min-batch size of the proposed method is 16 and epoch is set to 6. During the testing procedure, we employ the Generator $G$ for VeRi-776 $\to$ VehicleID translation and the Generator $F$ for VehicleID $\to$ VeRi-776 translation. The translated images are utilized for training reID models. For ATTNet, We implement the proposed vehicle re-id model in the Matconvnet \cite{VedaldiA} framework. We utilize stochastic gradient descent with a momentum of $\mu=0.0005$ during the training procedure. The batch size is set to 16. The learning rate of the first 50 epoch is set to 0.1, and the last 5 to 0.01.

\subsection{Evaluation}

\subsubsection{Comparison Methods}
There are really little methods about the vehicle reID of cross domain. So in this paper, we only discuss several methods and test them on VeRi-776 and VehicleID. Direct Transfer means directly applying the model trained by images from source domain on the target domain. CycleGAN \cite{KimT}, SPGAN \cite{DengW} and VTGAN are employed to translate images from source domain to target domain, and then the generated images are utilized to train reID model. Baseline \cite{ZhengZ'} denotes the compared training network of reID. ATTNet is our proposed network.

\subsubsection{Comparison of generated images}
\begin{figure}[ht]
\centering
\includegraphics[width=6.5cm]{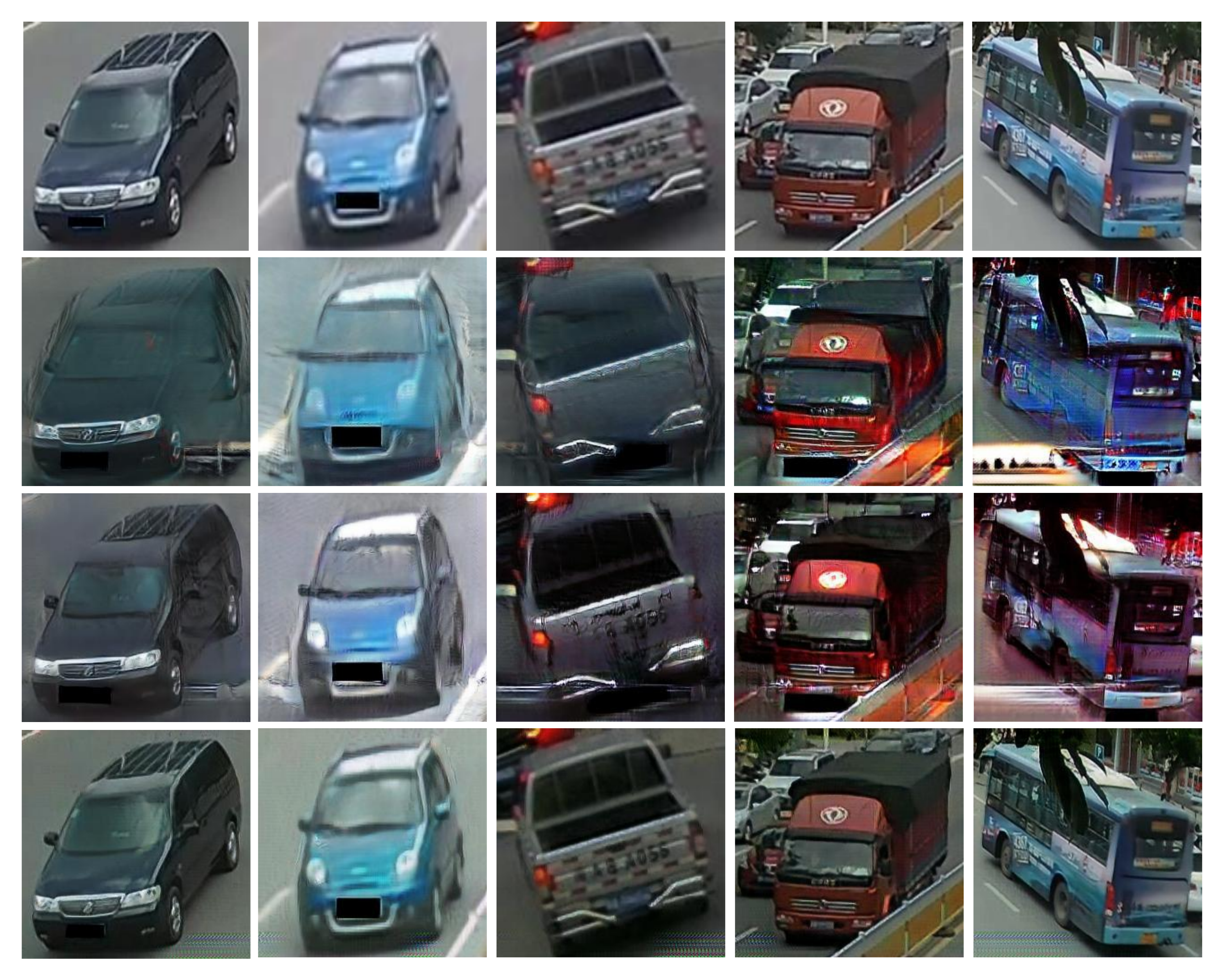}
\caption{The effect of the generated images. The first row is original images. The generated images using CycleGAN, SPGAN lie in the second row and third row respectively. The last row are generated images by VTGAN.} \label{fig5}
\end{figure}

To demonstrate the effectiveness of our proposed style transform model, the VehicleID and VeRi-776 are utilized to train the VTGAN. And CycleGAN and SPGAN are taken as compared methods. Fig.\ref{fig5} is the comparison results, which the source domain is VeRi-776, and target domain is VehicleID. For each group, the first row is the original images in VeRi-776. The second and third rows are generated by CycleGAN and SPGAN, respectively. The last row is generated by the proposed VTGAN.

From the Fig.5, we could find that, most images generated by CycleGAN are distorted seriously when transfer images from VeRi-776 to VehicleID. And though the SPGAN works better than the CycleGAN, the generated images also have evident deformation. However, for VTGAN, not only is the vehicle color and type information completely preserved, but also learns the style of the target dataset. As we can see from Fig.5, generated images by VTGAN have higher resolution and become darker, which learns from VehicleID.

\subsubsection{The impact of Image-Image Translation}

Firstly, we utilize CycleGAN to translate labeled images from the source domain to the target domain then train the baseline reID model with translated images in a supervised way. As shown in Table.\ref{tab1}, when trained on VeRi-776 training set using the baseline method and tested on VehicleID different testing sets, rank-1 accuracy improves from 35\% to 39.39\%, 30.42\% to 32.97\%, 27.28\% to 28.44\% and 25.41\% to 26.38\%, respectively. Through such an image-level domain adaptation method, effective domain adaptation baselines can be learned. From the Fig. \ref{fig5}, we could find that, though some of generated images by CycleGAN are distorted, the performance of reID model trained by generated images is improved. This illustrates methods of image-image translation have learned the important style information from the target domain, which could narrow-down the domain gap to a certain degree.

\subsubsection{The impact of VTGAN}

To verify the effectiveness of the proposed VTGAN, we conduct several experiments which training sets are images generated from different image translation methods. As shown in Table.\ref{tab1}, on VehicleID, compared with $CycleGAN+Baseline$, the gains of $VTGAN+Baseline$ are 5.05\%, 6\%, 6.66\% and 5.79\% in rank-1 of different test sets, respectively. Though SPGAN has better performance in the stage image-to-image translation than CycleGAN, it also causes deformation and color distortion in real scenario (see Fig.5). Hence, compared with $SPGAN+Baseline$, for different size of test sets on VehicleID, $VTGAN+Baseline$ has 1.57\%, 1.51\%, 1.56\% and 1.72\% improvements in mAP, respectively. All of these could demonstrate that the structure of VTGAN is more stable and could generate suitable samples for training in the target domain.

\subsubsection{The impact of ATTNet}

\begin{figure}[ht]
\centerline{
\subfloat[Test size=800]{\includegraphics[width=1.7in,height=1.3in]{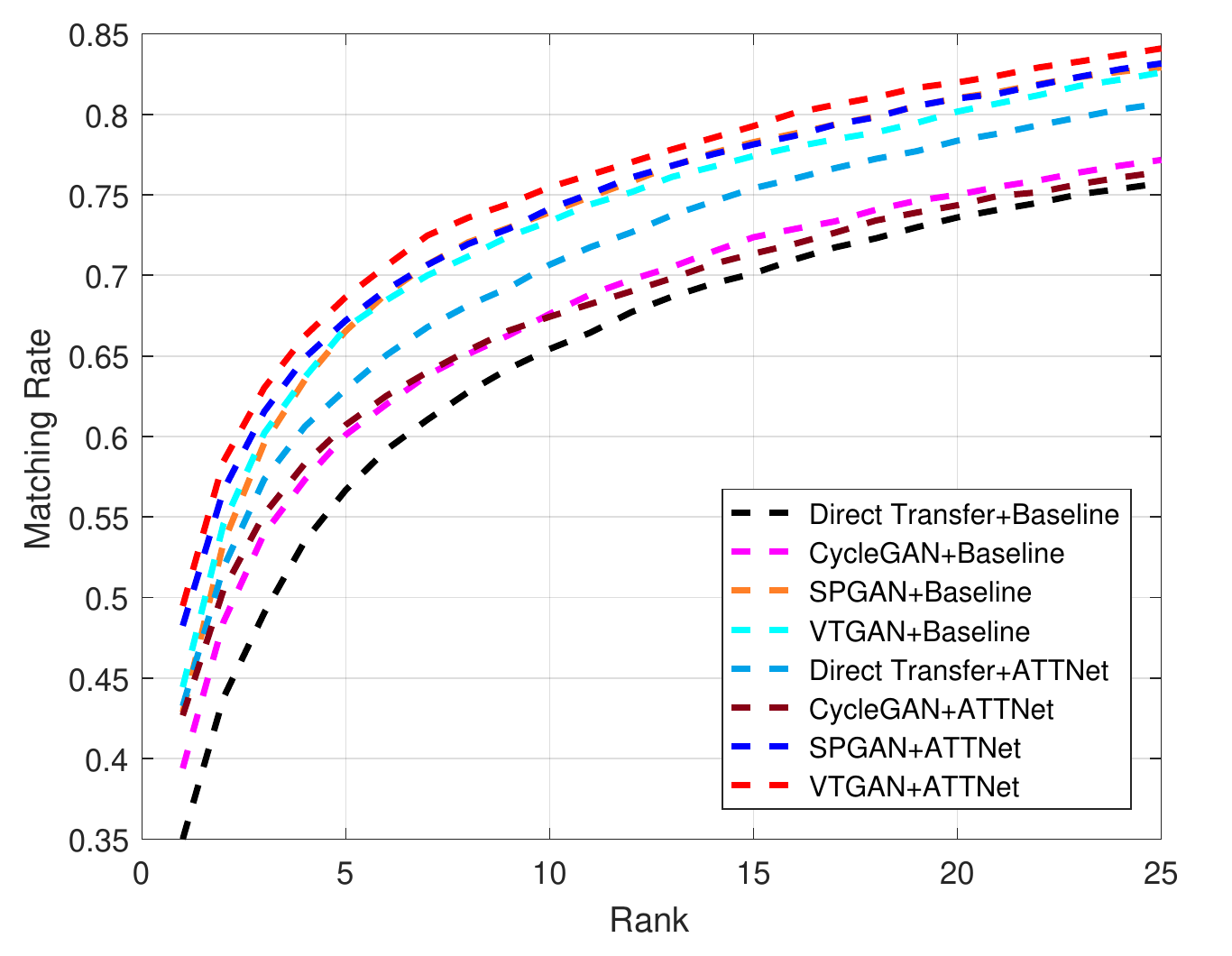}}
\subfloat[Test size=1600]{\includegraphics[width=1.71in,height=1.31in]{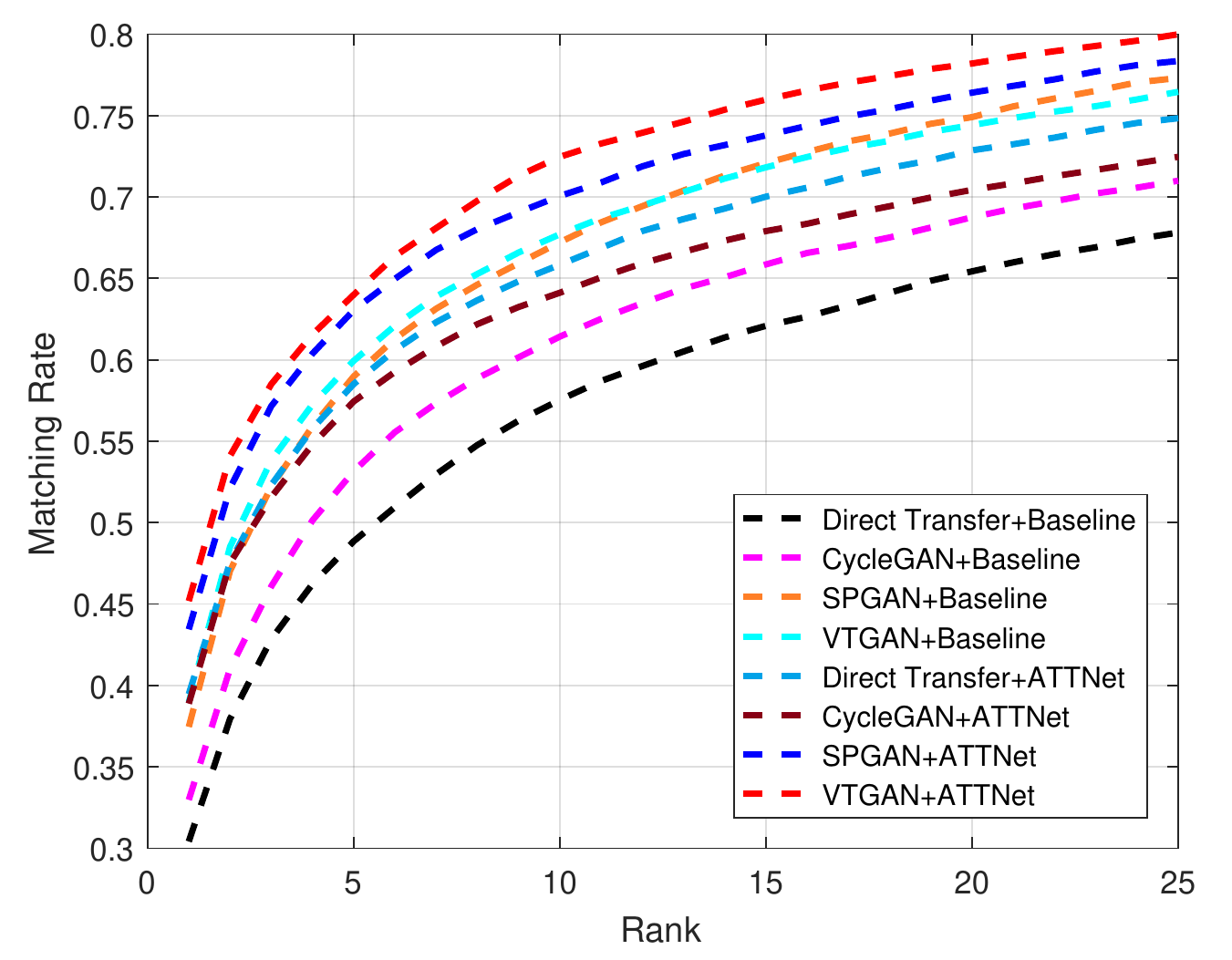}}
}
\centerline{
\subfloat[Test size=2400]{\includegraphics[width=1.7in,height=1.3in]{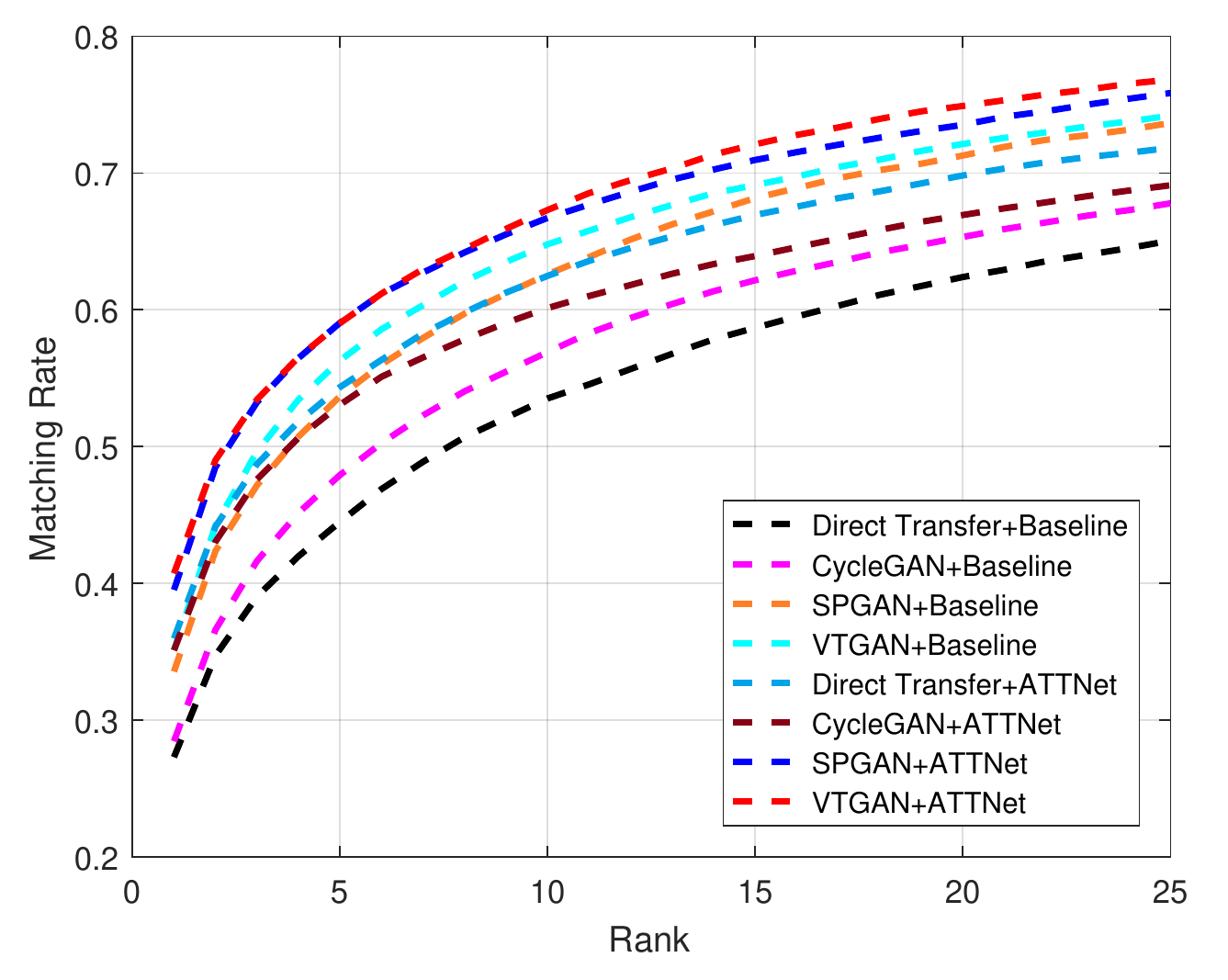}}
\subfloat[Test size=3200]{\includegraphics[width=1.7in,height=1.3in]{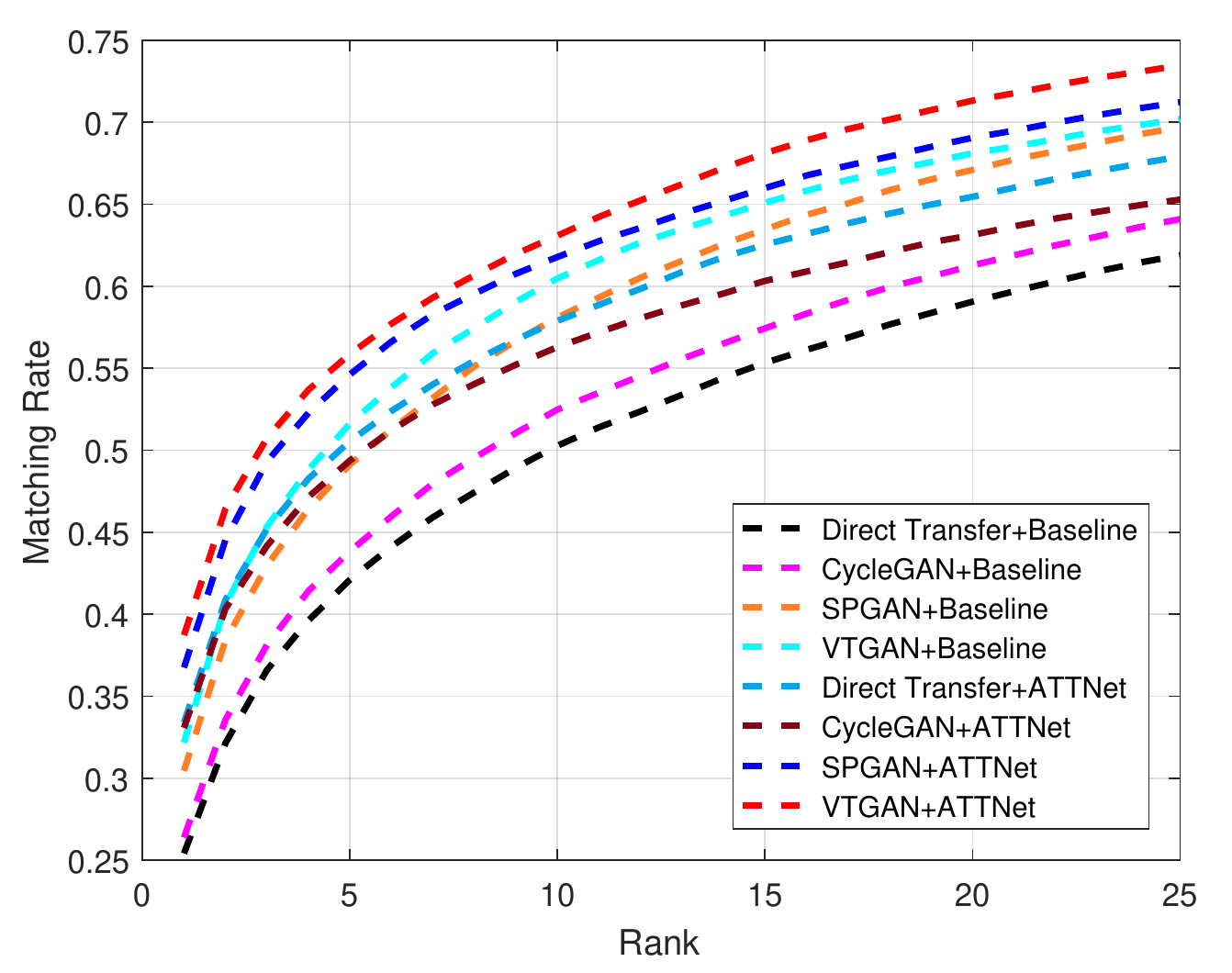}}
}
\caption{The CMC curves of different methods on VehicleID. (a) The result tested on the set with 800 vehicles. (b) The result tested on the set with 1600 vehicles. (c) The result tested on the set with 2400 vehicles. (d) The result tested on the set with 3200 vehicles.} \label{fig6}
\end{figure}

To further improve re-ID performance on target dataset, we propose ATTNet. Fig.\ref{fig6} is CMC resutls on VehicleID of different methods. As shown in Fig.\ref{fig6}, compared to methods with baseline reID model, not only original images but also generated images, methods using ATTNet have better performance. For instance, from the Table.\ref{tab1}, we could find that, compared with $Direct\ Transfer+Baseline$, $Direct \ Transfer+ATTNet$  has 8.26\%, 9.05\%, 8.67\%, and 7.99\% improvements in rank-1 of different test sets when the model is trained on VeRi-776 and tested on VehicleID. Besides, it is obvious that compared with the baseline methods, the reID model using the ATTNet have significant improvement for every image translation method. This demonstrates that the reID model trained by the proposed ATTNet can better adapt to cross-domain task than the baseline method.

\begin{table*}[htbp]
\center
\scriptsize
\setlength{\belowcaptionskip}{10pt}
\caption{Comparison of various domain adaptation methods over Baseline model and ATTNet-reID model on VehicleID. }\label{tab1}
\begin{tabular}{|C{2.6cm}|C{0.8cm}|C{0.8cm}|C{0.8cm}|C{0.8cm}|C{0.8cm}|C{0.8cm}|C{0.8cm}|C{0.8cm}|C{0.8cm}|C{0.8cm}|C{0.8cm}|C{0.8cm}|}
\hline
\multirow{2}*{Methods} & \multicolumn{3}{c|}{Test size = 800} & \multicolumn{3}{c|}{Test size = 1600}& \multicolumn{3}{c|}{Test size = 2400} & \multicolumn{3}{c|}{Test size = 3200} \\
\cline{2-13} & mAP(\%) & Rank1(\%) & Rank5(\%) & mAP(\%) & Rank1(\%) & Rank5(\%) & mAP(\%) & Rank1(\%) & Rank5(\%) & mAP(\%) & Rank1(\%) & Rank5(\%) \\

\hline
 Direct Transfer + Baseline    & 40.05	&35.00	&56.68	&34.90	&30.42	&48.85	&31.65	&27.28&	44.49 & 29.57 & 25.41 &42.11\\
 CycleGAN + Baseline   & 44.24	&39.39	&60.10	&37.68	&32.97	&53.16	&33.17	&28.44	&47.92 & 30.73 & 26.38 &43.84\\
 SPGAN + Baseline    &48.27	&42.87&	66.55	&42.51	&37.46	&58.97	&38.41	&33.54	&53.68 & 35.04 & 30.45 &49.13\\
 VTGAN + Baseline    & 49.53	&44.44	&66.74	&43.90	&38.97	&59.93	&40.07	&35.10	&56.29 & 36.86 & 32.17 & 51.63\\
\hline
 Direct Transfer + ATTNet    & 47.97	&43.26	&62.93	&43.94	&39.47	&58.51	&40.42	&35.95	&54.34 & 37.60 & 33.40 & 50.55\\
 CycleGAN + ATTNet    & 46.96	&42.68	&60.72	&43.27	&38.88	&57.44	&39.39	&35.09	&53.05 & 37.05 & 33.07 & 49.38\\
 SPGAN + ATTNet    & 52.72	&48.25	&67.20	&48.01	&43.44	&63.04	&44.17	&39.51	&59.05 & 41.05 & 36.75 & 54.63\\
 VTGAN + ATTNet    & 54.01	&49.48	&68.66	&49.72	&45.18	&63.99	&45.18	&40.71	&59.02 & 42.94 & 38.72 & 55.87\\
\hline
\end{tabular}
\center
\end{table*}

\section{Conclusion}
In this paper, we propose a vehicle reID framework based on GAN, which includes the VTGAN and ATTNet for domain adaptation. The VTGAN is designed to generate the vehicle images which preserve the identity information of source domain and learn the style of target domain. The ATTNet is proposed to train the reID model with generated images. It can be observed from the results that both the VTGAN and ATTNet achieve good results. What's more, it is obvious that existing datasets usually contain several viewpoints of vehicle images. It is also a limit for reID task in new domain. Hence, in our future studies, we would focus on using the GAN to generate the various viewpoints of vehicle images to expand the dataset and improve the performance of reID model.


\bibliographystyle{IEEEbib}
\bibliography{icme2019template}

\begin{thebibliography}{10}

\bibitem{wu20193D}
Lin Wu, Yang Wang, Ling Shao, and Meng Wang,
\newblock ``3-d personvlad: Learning deep global representations for
  video-based person reidentification,''
\newblock {\em IEEE transactions on neural networks and learning systems},
  2019.

\bibitem{wang2017effective}
Yang Wang, Xuemin Lin, Lin Wu, and Wenjie Zhang,
\newblock ``Effective multi-query expansions: Collaborative deep networks for
  robust landmark retrieval,''
\newblock {\em IEEE Transactions on Image Processing}, vol. 26, no. 3, pp.
  1393--1404, 2017.

\bibitem{Wu2018}
Lin Wu, Yang Wang, Junbin Gao, and Xue Li,
\newblock ``Where-and-when to look: Deep siamese attention networks for
  video-based person re-identification,''
\newblock {\em IEEE Transactions on Multimedia}, 2018.

\bibitem{wang2015robust}
Yang Wang, Xuemin Lin, Lin Wu, Wenjie Zhang, Qing Zhang, and Xiaodi Huang,
\newblock ``Robust subspace clustering for multi-view data by exploiting
  correlation consensus,''
\newblock {\em IEEE Transactions on Image Processing}, vol. 24, no. 11, pp.
  3939--3949, 2015.

\bibitem{wu2018deep}
Lin Wu, Yang Wang, Xue Li, and Junbin Gao,
\newblock ``Deep attention-based spatially recursive networks for fine-grained
  visual recognition,''
\newblock {\em IEEE Transactions on Cybernetics}, vol. 49, no. 5, pp.
  1791--1802, 2019.

\bibitem{DengW}
W.~Deng, L.~Zheng, and G~Kang,
\newblock ``Image-image domain adaptation with preserved self-similarity and
  domain-dissimilarity for person re-identification,''
\newblock {\em Proceedings of the IEEE Conference on Computer Vision and
  Pattern Recognition.}, pp. 994--1003, 2018.

\bibitem{WangJ}
J.~Wang, X.~Zhu, and S.~Gong,
\newblock ``Transferable joint attribute-identity deep learning for
  unsupervised person re-identification,''
\newblock {\em Proceedings of the IEEE Conference on Computer Vision and
  Pattern Recognition.}, pp. 2275--2284, 2018.

\bibitem{WeiL}
L.~Wei, S.~Zhang, and W.~Gao,
\newblock ``Person transfer gan to bridge domain gap for person
  re-identification,''
\newblock {\em Proceedings of the IEEE Conference on Computer Vision and
  Pattern Recognition.}, pp. 79--88, 2018.

\bibitem{wu2019cycle}
Lin Wu, Yang Wang, and Ling Shao,
\newblock ``Cycle-consistent deep generative hashing for cross-modal
  retrieval,''
\newblock {\em IEEE Transactions on Image Processing}, vol. 28, no. 4, pp.
  1602--1612, 2019.

\bibitem{Wang2016Iterative}
Yang Wang, Wenjie Zhang, Lin Wu, Xuemin Lin, Meng Fang, and Shirui Pan,
\newblock ``Iterative views agreement: An iterative low-rank based structured
  optimization method to multi-view spectral clustering,''
\newblock in {\em International Joint Conference on Artificial Intelligence},
  2016, pp. 2153--2159.

\bibitem{AMultiview}
Yang Wang, Lin Wu, Xuemin Lin, and Junbin Gao,
\newblock ``Multiview spectral clustering via structured low-rank matrix
  factorization,''
\newblock {\em IEEE transactions on neural networks and learning systems}, vol.
  29, no. 10, pp. 4833--4843, 2018.

\bibitem{KimT}
Kim T, Cha M, and Kim H,
\newblock ``Learning to discover cross-domain relations with generative
  adversarial networks,''
\newblock {\em Proceedings of the 34th International Conference on Machine
  Learning-Volume 70. JMLR. org.}, pp. 1857--1865, 2017.

\bibitem{ZhaoH}
Zhao H, Shi J, and Qi~X,
\newblock ``Pyramid scene parsing network,''
\newblock {\em IEEE Conf. on Computer Vision and Pattern Recognition}, pp.
  2881--2890, 2017.

\bibitem{TaigmanY}
Y.~Taigman, A.~Polyak, and L~Wolf,
\newblock ``Unsupervised cross-domain image generation,''
\newblock {\em arXiv preprint arXiv:1611.02200}, 2016.

\bibitem{HeK}
He~K, Zhang X, and Ren S,
\newblock ``Deep residual learning for image recognition,''
\newblock {\em Proceedings of the IEEE conference on computer vision and
  pattern recognition}, pp. 770--778, 2016.

\bibitem{LiuX'}
X.~Liu, W.~Liu, and T.~Mei,
\newblock ``Provid: Progressive and multimodal vehicle reidentification for
  large-scale urban surveillance,''
\newblock {\em IEEE Transactions on Multimedia}, vol. 20, no. 3, pp. 645--658,
  2018.

\bibitem{LiuH}
H.~Liu, Y.~Tian, and Y.~Yang,
\newblock ``Deep relative distance learning: Tell the difference between
  similar vehicles,''
\newblock {\em In Proceedings of the IEEE Conference on Computer Vision and
  Pattern Recognition}, pp. 2167--2175, 2016.

\bibitem{AbadiM}
M.~Abadi, P.~Barham, and J.~Chen,
\newblock ``Tensorflow: A system for large-scale machine learning,''
\newblock {\em In OSDI}, vol. 16, pp. 265--283, 2016.

\bibitem{VedaldiA}
Vedaldi A and Lenc K,
\newblock ``Matconvnet: Convolutional neural networks for matlab,''
\newblock {\em Proceedings of the 23rd ACM international conference on
  Multimedia. ACM}, pp. 689--692, 2015.

\bibitem{ZhengZ'}
Z.~Zheng, L.~Zheng, and Y.~Yang,
\newblock ``A discriminatively learned cnn embedding for person
  reidentification,''
\newblock {\em ACM Transactions on Multimedia Computing, Communications, and
  Applications (TOMM)}, vol. 14, no. 1, 2017.

\end{thebibliography}

\end{document}